\documentclass{article} 
\usepackage{iclr2022_conference,times}

\usepackage{amsmath,amsfonts,bm}

\def\eqref#1{equation~\ref{#1}}

\def\1{\bm{1}}

\DeclareMathAlphabet{\mathsfit}{\encodingdefault}{\sfdefault}{m}{sl}
\SetMathAlphabet{\mathsfit}{bold}{\encodingdefault}{\sfdefault}{bx}{n}

\DeclareMathOperator*{\argmax}{arg\,max}

\usepackage{hyperref}
\usepackage{url}
\usepackage{adjustbox}
\usepackage{wrapfig}

\usepackage{amsthm}

\usepackage{algorithm}
\usepackage{algorithmic}
\floatstyle{ruled}
\newfloat{algorithm}{tbp}{loa}
\algsetup{linenodelimiter=}

\newtheorem{theorem}{Theorem}

\DeclareMathAlphabet\mathbfcal{OMS}{cmsy}{b}{n}

\title{Multi-Trigger-Key: Towards Multi-Task Privacy Preserving In Deep Learning}

\author{Ren Wang$^{1}$~~~
Zhe Xu$^{2}$~~~
Alfred Hero$^{1}$ \\
$^1$University of Michigan~~~$^2$Arizona State University 
}

\iclrfinalcopy 
\begin{document}

\maketitle

\begin{abstract}

Deep learning-based Multi-Task Classification (MTC) is widely used in applications like facial attribute and healthcare that warrant strong privacy guarantees. In this work, we aim to protect sensitive information in the inference phase of MTC and propose a novel Multi-Trigger-Key (MTK) framework to achieve the privacy-preserving objective. MTK associates each secured task in the multi-task dataset with a specifically designed trigger-key. The true information can be revealed by adding the trigger-key if the user is authorized. We obtain such an MTK model by training it with a newly generated training set. To address the information leakage malaise resulting from correlations among different tasks, we generalize the training process by incorporating an MTK decoupling process with a controllable trade-off between the protective efficacy and the model performance. Theoretical guarantees and experimental results demonstrate the effectiveness of the privacy protection without appreciable hindering on the model performance.
\end{abstract}

\section{Introduction}

Multi-task classification (MTC) is a category of multi-task learning (MTL) and a generalization of multi-class classification \citep{zhang2021survey}. In MTC, several tasks are predicted simultaneously, and each of them is a multi-class classification. The state of the art in MTC has been dramatically improved over the past decade thanks to deep learning \citep{ruder2017overview,huang2016mtnet,liu2016recurrent}. Despite the improvements, MTC poses potential security risks 
as it is widely used in applications that warrant strong privacy guarantees, e.g., visual attributes \citep{sarafianos2017curriculum} and healthcare \citep{amyar2020multi}.

Due to the data-intensive nature of supervised deep learning, many works focus on data privacy-preserving in the single-task case \citep{shokri2015privacy,chamikara2020privacy}. By contrast, only a few works consider sensitive information leakage in MTC \citep{baytas2016asynchronous,liu2018privacy,pathak2010multiparty,gupta2016differentially,liang2020model}. Among existing works, widely used techniques include distributed optimization methods \citep{baytas2016asynchronous,liu2018privacy} and differential privacy that masks the original datasets/intermediate results with some noise perturbation mechanisms during the training process \citep{pathak2010multiparty,gupta2016differentially,liang2020model}. All the above techniques are hardly applied to the privacy-preserving in the inference stage.



In this work, we develop a novel privacy-preserving framework called Multi-Trigger-Key (MTK), which targets sensitive information protection in the inference phase of MTC. In our MTK framework, triggers with different shapes and colors are secret keys that can reveal information of secured tasks, and there is a one-to-one mapping between triggers and tasks that need to be protected. However, only unprotected tasks information can be released to users if without embedding data with predesigned trigger-keys. Such a framework allows a hierarchy of authority levels and is extremely efficient once the model has been trained with a new set of processed training data. Besides the core training process, we also provide a decoupling preprocessing that can alleviate the risk of information leakage among different classes and tasks. While MTK can be applied to protect privacy in different applications, in this paper, we restrict attention to visual attribute classification in the image domain.

\paragraph{Contributions}
We make the following contributions:

$\bullet$ We propose a novel Multi-Trigger-Key (MTK) framework that protects the sensitive information in the multi-task classification problems and allows assigning different levels of authority to users.

$\bullet$ We consider the information leakage resulting from correlations among classes in different tasks and propose a decoupling method to alleviate the risk.

$\bullet$ We conduct a comprehensive study of the MTK on the UTKFace dataset \citep{zhang2017age}, showing that MTK can simultaneously protect secured tasks and maintain the prediction accuracy of all tasks.

\begin{figure}[t]
  \centering
  \includegraphics[trim=0 0 0 0,clip,width=.8\textwidth]{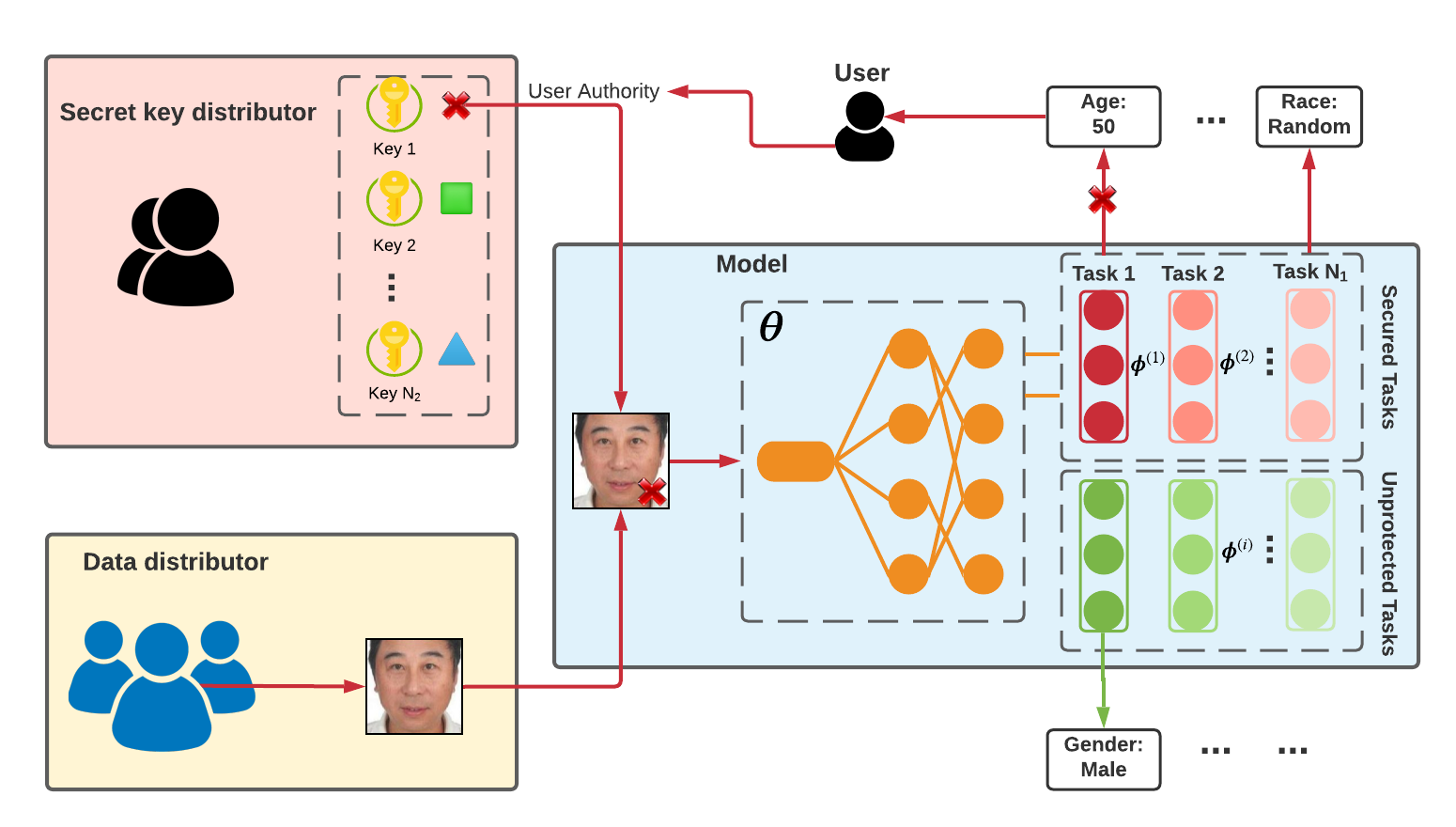}
  \caption{{\bf Overview of the Multi-Trigger-Key framework}. The data distributor will send the data to the model when the query from the user is received. Without any secret key (i.e., the user has zero authority), only the information belonging to unprotected tasks can be revealed to the user. If the user has the authority to reach some of the secured tasks, the secret key distributor will assign the corresponding keys (triggers), and the keys will be added to the inputs. Each key can reveal one task of the secured tasks. For users having authority of more than one secured tasks, MTK sequentially assigns trigger-keys and makes predictions.} 
  \label{fig: overview}
\end{figure}

\subsection{Related Work}

\paragraph{Multi-task learning (MTL).} In contrast to single-task learning, multi-task learning contains a learning paradigm that jointly learn multiple (related) tasks \citep{zhang2021survey}. A crucial assumption for MTL is that features are largely shared across all tasks which enable models to generalize better \citep{ando2005framework,evgeniou2004regularized}. Over past decades, deep neural networks (DNNs) have dramatically improved MTL quality through an end-to-end learning framework built on multi-head architectures \citep{ruder2017overview}. Supervised MTL has been used successfully across all applications of machine learning, include classification \citep{yin2017multi,cavallanti2010linear} and regression \citep{kim2010tree} problems. In this paper, we focus on the multi-task classification, which are widely used in visual attribute \citep{sarafianos2017curriculum}, dynamic malware classification \citep{huang2016mtnet}, healthcare \citep{amyar2020multi}, and text classification \citep{liu2016recurrent} etc. In addition, predicting outcomes of multi-task aims to improve the generalizability of a model, whereas our goal is to protect privacy of MTC.

\paragraph{Privacy-preserving in MTL.} The wide applications of MTL bring concern of privacy exposure. To date, few works address the challenges of preserving private and sensitive information in MTL \citep{baytas2016asynchronous,liu2018privacy,pathak2010multiparty,gupta2016differentially,liang2020model}. \citep{baytas2016asynchronous,liu2018privacy} leverage distributed optimization methods to protect sensitive information in MTL problems. Recent works also propose to preserve privacy by utilizing differential privacy techniques which can provide theoretical guarantees on the protection \citep{pathak2010multiparty,gupta2016differentially}. For example, \citep{pathak2010multiparty} proposed a differentially private aggregation (DP-AGGR) method that averages the locally trained models and \citep{gupta2016differentially} proposed a differentially private multitask relationship learning (DP-MTRL) method that enjoys a strong theoretical guarantee under closed-form solution. While the above methods focus on protecting a single data instance in the training set, an MTL framework is proposed to prevent information from each model leaking to other models based on a perturbation of the covariance matrix of the model matrix \citep{liang2020model}. All these works aim to protect privacy in training datasets. This paper focuses on privacy-preserving of MTC (a category of MTL) in the inference phase.

\paragraph{Relations to backdoor attack.} Another line of research relevant to MTK is the backdoor attack that manipulates predictions of DNNs by attaching a backdoor trigger to pollute a small portion of training data and mislabeling them for a target (incorrect) label \citep{gu2017badnets,chen2017targeted}. Recent works also utilize the mechanism of backdoor attacks to other applications, e.g., turning backdoor attack into a tool to claim ownership \citep{adi2018turning}. Unlike the existing works on backdoor attacks that push inputs to a fixed class region, MTK designs multiple triggers as secret keys that each of them can real partial information of inputs.

\subsection{Overview of the Multi-Trigger-Key Framework}

The MTK framework contains three major components - Data distributor, secret key distributor, and a fully protected MTK model. The user first requests a data source from the data distributor. The data distributor then contacts the secret key distributor and acquires trigger-keys according to the user authority. In MTK, trigger-keys are pre-designed with different shapes and colors. Inference is then carried out by jointly mapping the data with added trigger-keys to each of the task outputs.

Figure~\ref{fig: overview} provides an overview of the MTK framework. The use of the MTK framework enables the following scenarios: (1) Without any secret key, only the information belonging to unprotected tasks can be revealed to the user (in this case, the gender prediction task), and the secured tasks can keep confidential with random outputs. (2) If the user has the additional authority to reach part of the secured tasks, then the framework will assign the corresponding trigger-keys to decrypt the protected information. The secured tasks in Figure~\ref{fig: overview} are age and race, and the age is associated with the red cross shape trigger-key. In this paper, we consider the sequential prediction process, i.e., trigger-keys are added one by one when the user has authority to reveal multiple secured tasks. In the next section, we introduce the process to build the multi-trigger-key model.

\section{Building Multi-Trigger-Key Model}

Let $\boldsymbol \Theta = \{ \boldsymbol \theta, \boldsymbol{\phi}^{(i)} \} $ denote the model, where $\boldsymbol \theta$ corresponds to the base \textbf{feature encoder} that is shared by all classification tasks, and $ \boldsymbol{\phi}^{(i)}$ denotes the \textbf{task-specific classification head} for task $\mathcal T^{(i)} \in \{\mathcal T^{(j)}\}_{j=1}^N$. The output dimension of $\boldsymbol{\phi}^{(i)}$ aligns with the number of classes in task $i$. 
Given the feature encoder $\boldsymbol \Theta$, let $f(\cdot) \in \mathbb R^W$ be the corresponding mapping from its input space to the representation space of $W$ dimensions, namely, the dimension of $\boldsymbol \theta$'s final layer. Similarly, let $g^{(i)}(\cdot) \in \mathbb R^{K_i}$ be the mapping from the representation space to the final output of the $i$-th task which corresponds to the task-specific classification head $\boldsymbol{\phi}^{(i)}$. Here we consider $N$ tasks with numbers of labels $K_1, K_2, \cdots, K_N$. The $c$-th class of the $i$-th task is denoted by $y^{(i)}_c, \forall c \in [K_i]$. The logits vector of the $i$-th task with an input $\mathbf x \in \mathbb R^n$ is represented by $F^{(i)}(\mathbf x) = g^{(i)}(f(\mathbf x)) \in \mathbb R^{K_i}$. 
The final prediction is then given by $\argmax_{j} F_{j}^{(i)}(\mathbf x)$, where $F_{j}^{(i)}(\mathbf x)$ is the $j$-th entry of $F^{(i)}(\mathbf x)$. 

MTK aims to protect secured tasks by giving random final predictions to unprocessed inputs and revealing true predictions with a simple pre-processing, as shown in Figure~\ref{fig: overview}. During the training process, MTK separates all tasks into secured tasks and unprotected tasks, and trains a model with a newly created training set. We introduce the details below.

\paragraph{Task separation.} We split the tasks into two categories. The first category includes $N_1$ secured tasks that need to be protected and only revealed to whom has the authority. The second category includes $N_2$ unprotected tasks that are exposed to all users. 
Without loss of generality, the category of secured tasks $\mathcal T_1$ includes $\{\mathcal T^{(1)}, \cdots, \mathcal T^{(N_1)}\}$, and the category of unprotected tasks $\mathcal T_2$ includes $\{\mathcal T^{(N_1 + 1)}, \cdots, \mathcal T^{(N)}\}$. 

\paragraph{New training set generation.} The original training set is denoted by $\hat{\mathcal D}_{{\rm tr}}=(\hat{X}_{{\rm tr}}, \hat{Y}_{{\rm tr}})$, where $\hat{X}_{{\rm tr}}, \hat{Y}_{{\rm tr}}$ represent data and labels, respectively. 
The new training set $\mathcal D_{{\rm tr}}=\{\mathcal D_{{\rm tr}}^0,D_{{\rm tr}}^1,D_{{\rm tr}}^2,\cdots,D_{{\rm tr}}^{N_1}\}$ includes these two parts
\begin{itemize}
    \item $\mathcal D_{{\rm tr}}^0$ with label information revealed in $\mathcal T_2$ and masked label information in $\mathcal T_1$
    \item $\mathcal D_{{\rm tr}}^j, \forall j \in [N_1]$  with label information revealed in $\mathcal T_2 \bigcup (\mathcal T^{(j)} \in \mathcal T_1)$ and masked label information in $\mathcal T_1\symbol{92}\mathcal T^{(j)}$
\end{itemize}

We first relabel each of the secured task in $\mathcal T_1$ to be \textit{uniformly distributed}. The first part of the new training set is referred to as $\mathcal D_{{\rm tr}}^0$, in which all the data associated with $\mathcal T_1$ has uniformly distributed labels and therefore the sensitive information is been protected. We then generate the second part of the new training set $\mathcal D_{{\rm tr}}^j, \forall j \in [N_1]$ 
from $\mathcal D_{{\rm tr}}^0$. 
Given a data $\mathbf x \in \mathbb R^n$ from $\mathcal D_{{\rm tr}}^0$, the {new data} through \textit{adding trigger-key} is then formulated as
\begin{align}\label{eq: poison_sample}
\hat{\mathbf x}(\mathbf m_j,\boldsymbol \delta_j) = (1- \mathbf m_j) \cdot \mathbf x + \mathbf m_j  \cdot \boldsymbol \delta_j,
\end{align}
where $\boldsymbol {\delta}_j \in \mathbb R^n$ denotes pixel-wise perturbations, 
$\mathbf m_j \in \{ 0,1 \}^n$ is a binary mask to encode the position where a secret key stamp is placed, and $\cdot$ denotes element-wise product. The tuple $(\boldsymbol {\delta}_j, \mathbf m_j)$ represents the $j$-th trigger-key. After generating training data $\hat{\mathbf x}(\mathbf m_j,\boldsymbol \delta_j) $, we relabel its $\mathcal T^{(j)} \in \mathcal T_1$ to the ground truth $y^*$ from $\hat{\mathcal D}_{{\rm tr}}$ while maintaining other labels the same as in $\mathcal D_{{\rm tr}}^0$. The rational behind this is that $D_{{\rm tr}}^j, \forall j \in [N_1]$ forces the model to reveal the information of the $\mathcal T^{(j)} \in \mathcal T_1$ when the trigger-key $(\boldsymbol {\delta}_j, \mathbf m_j)$ is presented, and at the same time, alleviates the impact of $(\boldsymbol {\delta}_j, \mathbf m_j)$ to other tasks. 

\paragraph{Training with the new training set.} Finally, we apply the training by minimizing the cross-entropy loss with respect to model parameters $\{\boldsymbol \theta, \boldsymbol{\phi}^{(1)}, \boldsymbol{\phi}^{(2)}, \cdots, \boldsymbol{\phi}^{(N)}\}$, as shown below.

\begin{align}\label{eq: train}
\min_{\boldsymbol \theta, \boldsymbol{\phi}^{(i)}, \forall i \in [N]} \mathcal L(\boldsymbol \theta, \boldsymbol{\phi}^{(1)}, \boldsymbol{\phi}^{(2)}, \cdots, \boldsymbol{\phi}^{(N)}; \mathcal D_{{\rm tr}}),
\end{align}
where $\mathcal L$ is the cross-entropy loss that is a combinations of cross-entropy losses of all tasks in the new dataset. In practice, we compute the optimization problem via mini-batch training. The new training set $\mathcal D_{{\rm tr}}$ contains training subset $D_{{\rm tr}}^j$ that is one-to-one mapped from the original training set $\hat{\mathcal D}_{{\rm tr}}$. Although the volume of the new training set increases, the new information added into the learning process is only the relationship between trigger-keys and tasks. Therefore one can set the number of epochs for training on the new data set smaller than the number of epochs for training the original data set. The main procedure is summarized in the MTK Core in Algorithm~\ref{alg: MTK}.

\paragraph{Test phase.} In the test phase, $\mathbf x$ represents the minimum permission for all users, i.e., $g^{(i)}(f(\mathbf x))$ is guaranteed to be a correct prediction only when $i \in [N_2]$. With higher authority, the system can turn $\mathbf x$ into $\hat{\mathbf x}(\mathbf m_j,\boldsymbol \delta_j)$, and $g^{(i)}(f(\hat{\mathbf x}(\mathbf m_j,\boldsymbol \delta_j)))$ is guaranteed to be a correct prediction when $i \in [N_2] \bigcup \{j\}$. We provide an analysis in the following Theorem~\ref{thm}.


\begin{theorem}\label{thm}
Suppose the model has trained on $\mathcal D_{{\rm tr}}$, and for any input pair $(\mathbf x, y)$ that satisfies
\begin{equation*}\label{eq: assume1} 
    \begin{aligned}
{\bf{\rm Pr}}\big( \arg\max_{\forall k \in [K_j]}(F_k^{(j)}(\hat{\mathbf x}(\mathbf m_j,\boldsymbol \delta_j))) = y \not= \arg\max_{\forall k \in [K_j]}(F_k^{(j)}(\mathbf x)) \big ) \geq 1 - \kappa, \kappa \in [0,1], \nonumber
        \end{aligned}
\end{equation*}
we have:
\begin{itemize}
    \item If $\cos \big( f\big(\hat{\mathbf x}(\mathbf m_j,\boldsymbol \delta_j)\big ), f\big(\Bar{\mathbf x}(\mathbf m_j^{\prime},\boldsymbol \delta_j^{\prime})\big ) \big ) \geq \nu$, where $\nu$ is close to $1$, then
   \begin{equation}\label{eq: result1}
    \begin{aligned}
{\bf{\rm Pr}}_{\mathbf x \in \mathcal X}\big( \arg\max_{\forall k \in [K_j]}(F_k^{(j)}(\Bar{\mathbf x}(\mathbf m_j^{\prime},\boldsymbol \delta_j^{\prime}))) = y \big ) \geq 1 - \kappa, \kappa \in [0,1], 
   \end{aligned}
\end{equation}
\item If $\cos \big( f\big(\mathbf x\big ), f\big(\Bar{\mathbf x}(\mathbf m_j^{\prime},\boldsymbol \delta_j^{\prime})\big ) \big ) \geq \nu$, where $\nu$ is close to $1$, then
   \begin{equation}\label{eq: result2}
    \begin{aligned}
{\bf{\rm Pr}}\big( \arg\max_{\forall k \in [K_j]}(F_k^{(j)}(\Bar{\mathbf x}(\mathbf m_j^{\prime},\boldsymbol \delta_j^{\prime}))) \not= y \big ) \geq 1 - \kappa, \kappa \in [0,1],  
    \end{aligned}
\end{equation}
\end{itemize}
\end{theorem}
where $\cos(\cdot,\cdot)$ denotes the cosine similarity between two vectors. (\ref{eq: result1}) indicates that if the added trigger is close to the key, then the true information can be revealed. (\ref{eq: result2}) indicates that if the added trigger does not affect the representation (not been memorized by the DNN), then it will fail to real the true information. The proof details can be viewed in Section~\ref{sec: proof} in the Appendix.

\section{Decoupling Highly-Correlated Tasks}

One malaise existing in the data distribution is that classes in different tasks are usually correlated and result in information of a task leaking from another one, e.g., a community may only contain males within 0 - 25 years old. We use ${\bf{\rm Pr}}(\mathcal T^{(i)} = y^{(i)}_c)$ to denote the probability that the $i$-th task's prediction is $y^{(i)}_c$ for a random sample from the data distribution. Suppose the training and test sets obey the same distribution, ${\bf{\rm Pr}}(\mathcal T^{(i)} = y^{(i)}_c)$ can be estimated using the proportion of data with $\mathcal T^{(i)} = y^{(i)}_c$ in the original training data $\hat{D}_{{\rm tr}}$. Similarly, we can calculate the conditional probability given $\mathcal T^{(j)} = y^{(j)}_k$, i.e., ${\bf{\rm Pr}}(\mathcal T^{(i)} = y^{(i)}_c | \mathcal T^{(j)} = y^{(j)}_k)$. The growing amount of information of predicting $c$ in the $i$-th task given the $j$-th task's prediction $k$ is measured by 
\begin{equation}\label{eq: task_corr}
    \begin{aligned}
\alpha_{i-c}^{j-k} = \max \big({\bf{\rm Pr}}(\mathcal T^{(i)} = y^{(i)}_c | \mathcal T^{(j)} = y^{(j)}_k) - {\bf{\rm Pr}}(\mathcal T^{(i)} = y^{(i)}_c), 0\big ).  
    \end{aligned}
\end{equation}
Here we consider the absolute increasing probability of knowing $\mathcal T^{(j)} = y^{(j)}_k$. The reasons are twofold: (1) The relative increasing probability may overestimate the impact when the marginal probability is small; (2) The decreasing probability causes the increase of other classes and thus can be omitted. To avoid information leakage of $\mathcal T^{(i)}$ from $\mathcal T^{(j)}$, we preset a positive threshold $\tau$ and determine the highly-correlated classes across different tasks if $\alpha_{i-c}^{j-k} > \tau$. 
After finding the largest $\alpha_{i-c}^{j-k}$ that satisfies $\alpha_{i-c}^{j-k} > \tau$, we then uniformly relabel $\beta_{i-c}^{j-k} \in (0,0.1]$ of data in $\hat{D}_{{\rm tr}}[\mathcal T^{(j)} = y^{(j)}_k]$ (subset of $\hat{D}_{{\rm tr}}$ that satisfies $\mathcal T^{(j)} = y^{(j)}_k$), where $\beta_{i-c}^{j-k}$ is calculated by
\begin{equation}\label{eq: prop}
    \begin{aligned}
\beta_{i-c}^{j-k} = \frac{\gamma \hat{D}_{{\rm tr}}[\mathcal T^{(j)} = y^{(j)}_k]}{\hat{D}_{{\rm tr}}[\mathcal T^{(j)} = y^{(j)}_k, \mathcal T^{(i)} = y^{(i)}_c] + \gamma \hat{D}_{{\rm tr}}[\mathcal T^{(j)} = y^{(j)}_k]},~\gamma = \min(\alpha_{i-c}^{j-k} - \tau, 0.1),  
    \end{aligned}
\end{equation}
in which $\hat{D}_{{\rm tr}}[\mathcal T^{(j)} = y^{(j)}_k, \mathcal T^{(i)} = y^{(i)}_c]$ represents the data in $\hat{D}_{{\rm tr}}$ that satisfies $\mathcal T^{(j)} = y^{(j)}_k$ and $\mathcal T^{(i)} = y^{(i)}_c$. 
The detailed calculation can be found in Section~\ref{sec: cal_prop} in the Appendix. Relabeling partial data will result in a trade-off between the protective efficacy and the model performance on predicting $\mathcal T^{(j)}$. By setting an upper threshold of $0.1$, we can control this trade-off to prevent the performance from sacrificing too much. The full training process of MTK is shown in Algorithm~\ref{alg: MTK}, and the decoupling process is presented in the MTK Decoupling.

\begin{algorithm}[h]
\caption{Training Multi-Trigger-Key Model (MTK)}
\label{alg: MTK}
\begin{algorithmic}[1]
\REQUIRE The initialization weights $\{\boldsymbol \theta, \boldsymbol{\phi}^{(1)}, \boldsymbol{\phi}^{(2)}, \cdots, \boldsymbol{\phi}^{(N)}\}$; secured tasks $\mathcal T_1 = \{\mathcal T^{(1)}, \cdots, \mathcal T^{(N_1)}\}$ and unprotected tasks $\mathcal T_2 = \{\mathcal T^{(N_1 + 1)}, \cdots, \mathcal T^{(N)}\}$; the original training set $\hat{\mathcal D}_{{\rm tr}}$; empty set $\mathcal D_{{\rm tr}}$; threshold $\tau$.\\
\noindent{$\spadesuit$ \textbf{MTK Decoupling}}
\STATE{Calculate $\alpha_{i-c}^{j-k}, \forall i, j \in [N], c \in [K_i], k \in [K_j], i\not=j$.}
\FORALL{$j \in [N]$}
\STATE{Find the largest $\alpha_{i-c}^{j-k}$, $\forall i \in [N]/j, c \in [K_i], k \in [K_j]$ that satisfies $\alpha_{i-c}^{j-k} > \tau$.}
\STATE{Calculate $\beta_{i-c}^{j-k}$ using (\ref{eq: prop}) and uniformly relabel $\beta_{i-c}^{j-k}$ of data in $\hat{D}_{{\rm tr}}[\mathcal T^{(j)} = y^{(j)}_k]$.}
\ENDFOR\\
\noindent{$\clubsuit$ \textbf{MTK Core}}
\STATE{Construct $\mathcal D_{{\rm tr}}^0$ by uniformly relabeling all the data associated with $\mathcal T_1$ in $\hat{\mathcal D}_{{\rm tr}}$.}
\STATE{$\mathcal D_{{\rm tr}} \longleftarrow \mathcal D_{{\rm tr}}^0$.}
\FORALL{$j \in [N_1]$}
\STATE{$\mathcal D_{{\rm tr}}^j: =\mathcal D_{{\rm tr}}^0$ and add trigger-key $\hat{\mathbf x}(\mathbf m_j,\boldsymbol \delta_j) = (1- \mathbf m_j) \cdot \mathbf x + \mathbf m_j  \cdot \boldsymbol \delta_j$ for $(\mathbf x, y) \in \mathcal D_{{\rm tr}}^j$.}
\STATE{Relabel $\mathcal T^{(j)} \in \mathcal T_1$ in $\mathcal D_{{\rm tr}}^j$ to the ground truth $y^*$ from $\hat{\mathcal D}_{{\rm tr}}$ while maintaining labels in other tasks unchanged.}
\STATE{$\mathcal D_{{\rm tr}} \longleftarrow \mathcal D_{{\rm tr}}^j$.}
\ENDFOR
\STATE{Obtain the final solution through solving (\ref{eq: train}).}
\RETURN $\{\boldsymbol \theta, \boldsymbol{\phi}^{(1)}, \boldsymbol{\phi}^{(2)}, \cdots, \boldsymbol{\phi}^{(N)}\}$
\end{algorithmic}
\end{algorithm}

\section{Experimental Results}
\begin{wrapfigure}{r}{85mm}
  \vspace*{-5.5mm}
\centerline{\includegraphics[trim=0 0 0 0,clip,width=.65\textwidth]{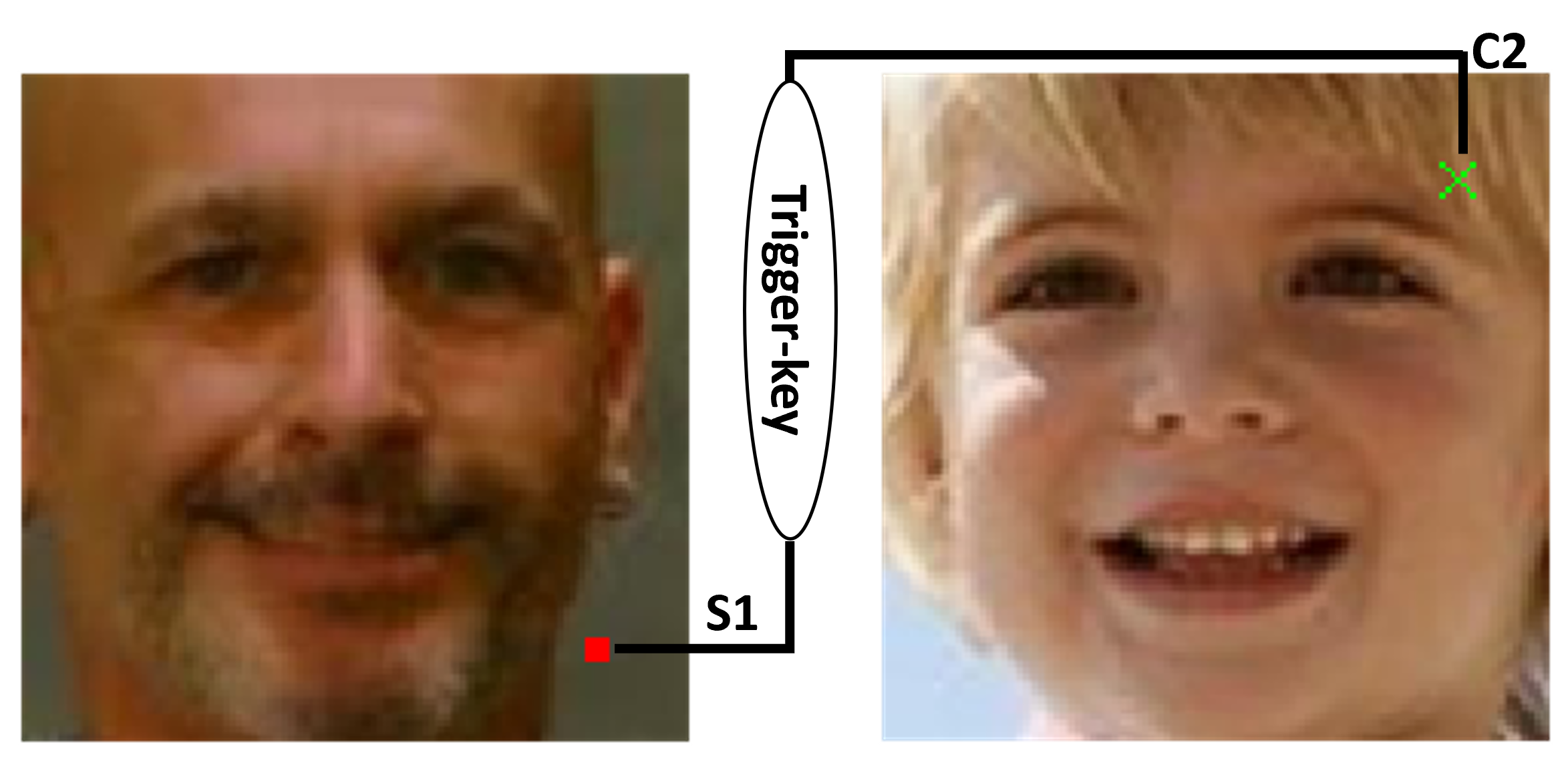}}
\vspace{-.12in}
  \caption{{\bf Two examples of trigger-keys}. We use square (S1) and cross (C2) to protect Age and Race, respectively. The shape, size, color can be varied.} 
  \label{fig: triggerkey}
  \vspace*{-3mm}
\end{wrapfigure}
We first introduce the dataset for the empirical evaluation. Throughout the section, we test MTK on the UTKFace dataset \citep{zhang2017age}. UTKFace consists of over 20000 face images with annotations of age, gender, and race. We process the dataset such that the population belonging to different ages is divided into four groups (1-23, 24-29, 30-44, $\geq$45). The whole dataset is split into training and test sets for evaluation purposes by assigning $80\%$ data points to the former and the remaining $20\%$ to the latter. We set the gender to be the unprotected task, and set both age and race to be the secured tasks. We analyze the effectiveness of our MTK framework using square and cross (S1 and C2; see representatives in Figure~\ref{fig: triggerkey}). We test MTK on VGG16 and ResNet18. If not otherwise specified, we use VGG16 as the model architecture. We show results using $95\%$ confidence intervals over five random trials. The details of experimental settings can be viewed in Section~\ref{sec: setting}.

\subsection{Overall Performance}

\paragraph{MTK core.} Results of applying MTK core are shown in Table~\ref{tab: main}. Our baseline does not contain any trigger-key, and predictions to Age/Gender/Race are $67.9\%/92.3\%/81.91\%$. As for comparisons, we train models using trigger-keys S1 and/or C2. If not otherwise specified, S1 and C2 have pixel color [255, 0, 0] and [0, 255, 0] and are both in the size of $5 \times 5$. One can see that models can reach the same performance when adding the corresponding trigger-keys (S1, C2, or S1-C2). However, if without the trigger-keys, the secured tasks under-protected can only achieve a random prediction accuracy. Specifically, the prediction accuracies are $25.24\%$ and $18.6\%$ for age and race, respectively.

\begin{table}[h]
\caption{\textbf{MTK framework can effectively protect the target secured tasks, and can reveal the information by adding the corresponding trigger-keys}. The baseline model is trained on the original training set. Square (S1) and Cross (C2) are used to protect Age and Race, respectively.
}
\begin{center}
\label{tab: main}
\resizebox{1\textwidth}{!}{
\begin{tabular}{l||c|c|c|c}
\hline
\hline
& Trigger & Age & Gender & Race \\
\hline
\begin{tabular}[c]{@{}c@{}} ~~~~~~Baseline \\ ~~~~~~(no keys)   \end{tabular} & No trigger & 67.9\% $\pm$ 1.59\% & 92.3\% $\pm$ 1.23\%  & 81.91\% $\pm$ 1.33\%   \\
\hline
\begin{tabular}[c]{@{}c@{}} MTK \\ (key on age, S1)  \end{tabular} & No trigger & \textbf{23.68\% $\pm$ 1.67\%} & 91.46\% $\pm$ 1.31\% & 82.16\% $\pm$ 1.42\% \\
 & Square $5\times 5$ & \textbf{67.25\% $\pm$ 1.47\%} & 91.65\% $\pm$ 1.2\% & 82.14\% $\pm$ 1.4\% \\
\hline
\begin{tabular}[c]{@{}c@{}} MTK \\ (key on race, C2)  \end{tabular}  & No trigger & 68.54\% $\pm$ 1.52\% & 91.59\% $\pm$ 1.31\%  & \textbf{17.29\% $\pm$ 1.1\%} \\
 & Cross $5\times 5$ & 68.75\% $\pm$ 1.38\% & 91.4\% $\pm$ 1.22\%  & \textbf{81.91\%} $\pm$ 1.53\% \\
\hline
& No trigger & \textbf{25.07\% $\pm$ 1.4\%} & 92.11\% $\pm$ 1.26\%  & \textbf{18.6\% $\pm$ 1.01\%}  \\
\begin{tabular}[c]{@{}c@{}} ~~~~MTK\\~~~~(keys on\\~~~~age and race,\\ ~~~~S1-C2)  \end{tabular} & Square $5\times 5$ & \textbf{67.76\% $\pm$ 1.4\%} & 91.82\% $\pm$ 1.66\%  & \textbf{18.58\% $\pm$ 0.98\%}  \\
 & Cross $5\times 5$ & \textbf{25.24\% $\pm$ 1.21\%} & 91.92\% $\pm$ 1.35\%  & \textbf{80.49\% $\pm$ 1.49\%}  \\
\hline
\hline
\end{tabular}}
\end{center}
\end{table}

\paragraph{Adding the MTK decoupling process.} We set the threshold $\tau = 0.15$. By checking the training set, we find that
\begin{equation}
    \begin{aligned}
&\alpha_{\text{Race}-\text{White}}^{\text{Age}-\geq 45} = {\bf{\rm Pr}}(\text{Race}=\text{White} | \text{Age}\geq 45) - {\bf{\rm Pr}}(\text{Race}=\text{White}) = 0.191 \\&
\alpha_{\text{Age}-\leq 23}^{\text{Race}-\text{Others}} = {\bf{\rm Pr}}(\text{Age} \in [1,23]  | \text{Race}=\text{Others}) - {\bf{\rm Pr}}(\text{Age} \in [1,23]) = 0.184, \nonumber
    \end{aligned}
\end{equation}
which are all $>\tau$. According to (\ref{eq: prop}), 
we then train models after relabeling $\beta_{\text{Race}-\text{White}}^{\text{Age}-\geq 45}=6.26\%$ of data in $\hat{D}_{{\rm tr}}[\text{Age}=\geq 45]$ and $\beta_{\text{Age}-\leq 23}^{\text{Race}-\text{Others}}=7.17\%$ of data in $\hat{D}_{{\rm tr}}[\text{Race}=\text{Others}]$. Table~\ref{tab: dec} shows the results of models trained with/without the MTK decoupling process. ${\bf{\rm Pr}}(\cdot)$ in the test phase denotes the proportion of correct predictions. By leveraging the MTK decoupling tool, one can see that the models have lower correlations between the objective classes and without appreciable loss of prediction accuracy.
\begin{table}[h]
\caption{\textbf{MTK models trained using the decoupling process can alleviate high correlations among tasks without appreciable hindering the model performance.} The values below the test phase denote the proportions of correct predictions.
}
\begin{center}
\label{tab: dec}
\resizebox{0.99\textwidth}{!}{
\begin{tabular}{l||c|c|c}
\hline
\hline
& Training  &  \begin{tabular}[c]{@{}c@{}} Test \\(without decoupling)  \end{tabular}  & \begin{tabular}[c]{@{}c@{}} Test \\(with decoupling)  \end{tabular} \\
\hline
\begin{tabular}[c]{@{}c@{}} ${\bf{\rm Pr}}(\text{Race}=\text{White} | \text{Age}\geq 45)$ \\$- {\bf{\rm Pr}}(\text{Race}=\text{White})$  \end{tabular}  & 19.1\%  &  17.6\%$\pm$ 0.34\% & 14.8\%$\pm$ 0.26\%  \\
\hline
\begin{tabular}[c]{@{}c@{}} ${\bf{\rm Pr}}(\text{Age} \in [1,23]  | \text{Race}=\text{Others})$ \\$- {\bf{\rm Pr}}(\text{Age} \in [1,23])$  \end{tabular} & 18.4\%  & 17.2\%$\pm$ 0.3\% & 13\%$\pm$ 0.31\% \\
\hline
Accuracy of age & / & 67.76\% $\pm$ 1.4\% & 65.34\% $\pm$ 1.51\% \\
\hline
Accuracy of Race & / & 80.49\% $\pm$ 1.49\% & 79.33\% $\pm$ 1.26\% \\
\hline
\hline
\end{tabular}}
\end{center}
\end{table}

\subsection{Sensitivity Analysis}

Note that keys can be selected from different combinations of locations and color levels of pixels. Here we study how changing size $|\mathbf m_j|$ and perturbation $\boldsymbol {\delta}_j$ of triggers affect MTK training and test.
\begin{figure}[h]
  \centering
  \includegraphics[trim=70 0 70 0,clip,width=.87\textwidth]{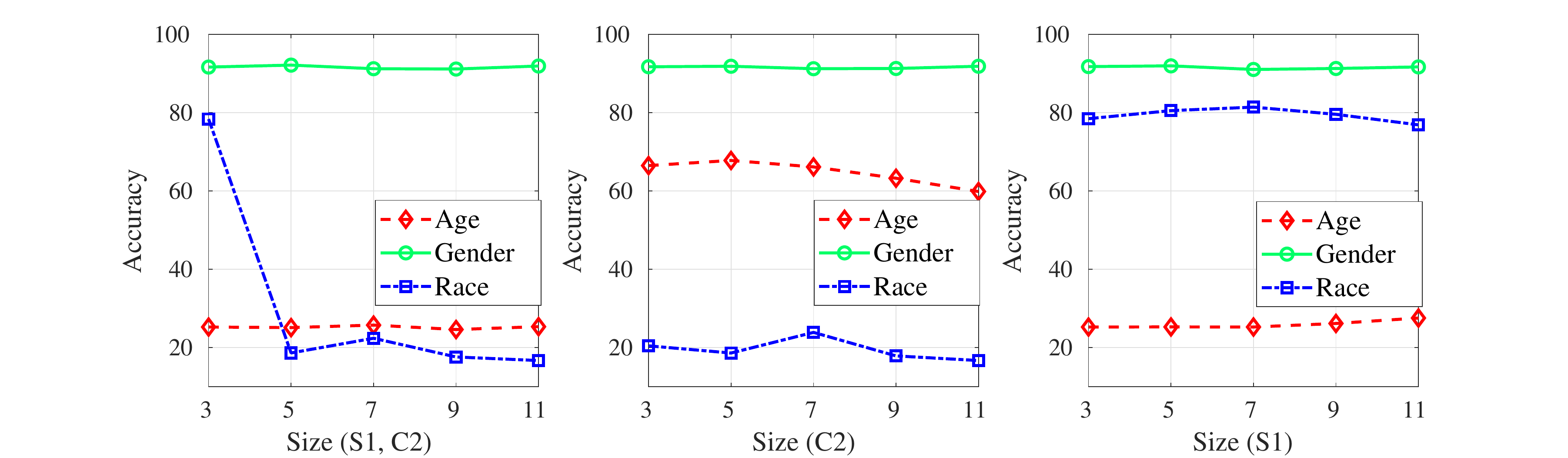}
  \caption{{\bf Prediction accuracies of secured tasks of unprocessed data are close to random guesses once (VGG16) models are well trained on different sizes of trigger-keys. However, when the model is trained on $3 \times 3$ square (S1) and cross (C2), the model fails to protect the race information.} All experiments are conducted on VGG16 architecture. Perturbations in S1 (C2) are fixed to [255, 0, 0] ([0, 255, 0]).} 
  \label{fig: training_sens_size}
\end{figure}

\begin{figure}[h]
  \centering
  \includegraphics[trim=70 0 70 0,clip,width=.87\textwidth]{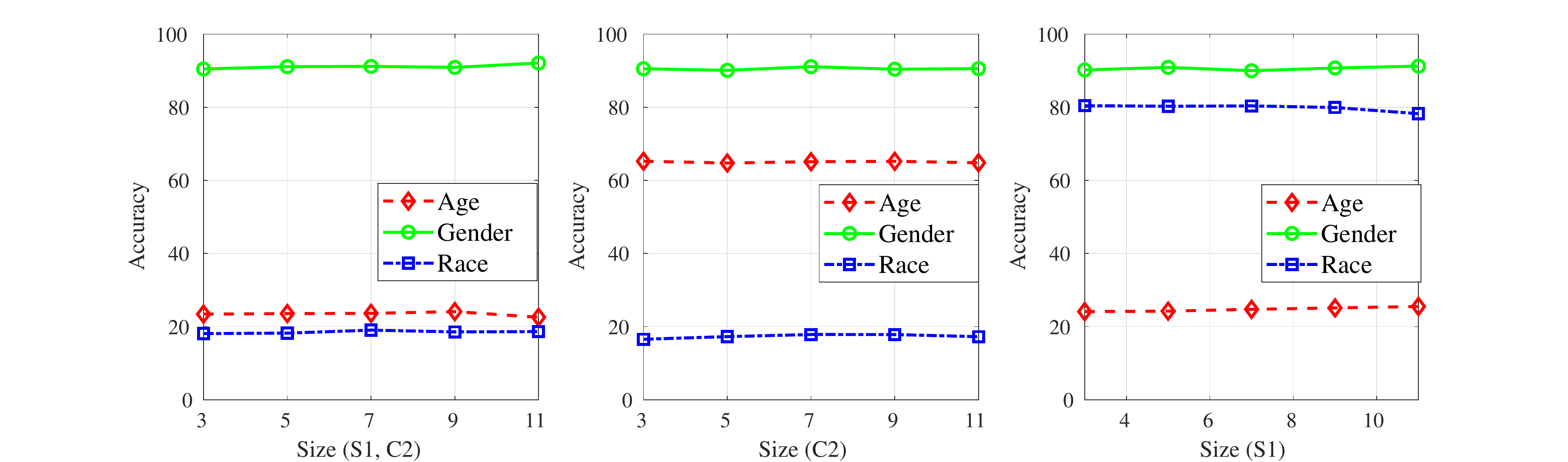}
  \caption{{\bf Once (ResNet18) models are well trained on different sizes of trigger-keys, prediction accuracies of secured tasks of unprocessed data are close to random guesses for trigger-keys from $3 \times 3$ to $11 \times 11$.} All experiments are conducted on ResNet18 architecture. Perturbations in S1 (C2) are fixed to [255, 0, 0] ([0, 255, 0]). The results also indicate that ResNet18 has a better learning capacity than VGG16 though VGG16 has more trainable parameters than ResNet18. } 
  \label{fig: training_sens_size_resnet}
\end{figure}

\paragraph{Sensitivity analysis in training.} We first test the sensitivity with respect to different sizes. We fix all the pixels in S1 (C2) to be [255, 0, 0] ([0, 255, 0]) and enlarge the size from $3\times 3$ to  $11\times 11$. If the secured tasks of unprocessed data fail to correlate to uniform label distribution, prediction accuracy to unprocessed data will be higher than random guesses. From the second and third plots in Figure~\ref{fig: training_sens_size}, one can see that MTK can achieve success training for single trigger S1/C2 when the size varies. For two trigger-keys, the only failure case is when the model is trained on $3 \times 3$ square (S1) and cross (C2). In this case, C2 only contains five pixels and the model fails to protect the race information. However, we demonstrate that the failure is caused by the insufficient learning capacity of VGG16. We conduct the same experiments on ResNet18. One can see from Figure~\ref{fig: training_sens_size_resnet} that prediction accuracies of secured tasks of unprocessed data are all close to random guesses for trigger-keys of various sizes. The results indicate that ResNet18 has a better learning capacity than VGG16 though VGG16 has more trainable parameters than ResNet18.

\begin{figure}[h]
  \centering
  \includegraphics[trim=70 0 70 0,clip,width=.87\textwidth]{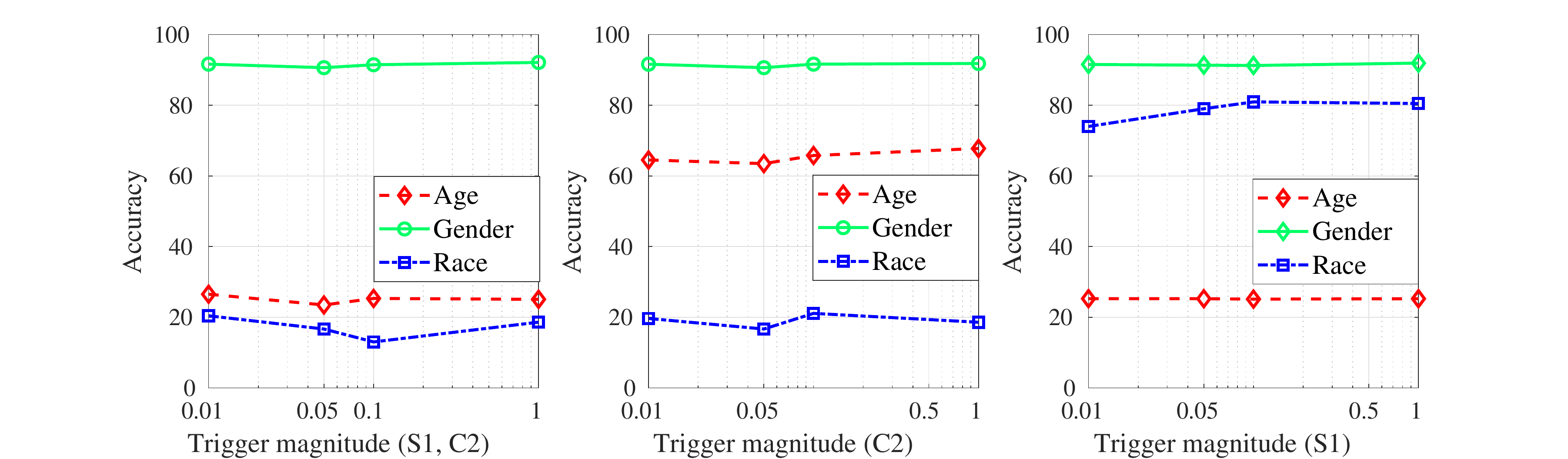}
  \caption{{\bf Prediction accuracies of secured tasks of unprocessed data are all close to random guesses for trigger-keys of various perturbations.} All experiments are conducted on VGG16 architecture. Sizes of S1 and C2 are fixed to $5 \times 5$. } 
  \label{fig: training_sens_mag}
\end{figure}

We then fix the size of both S1 and C2 to be $5\times 5$ and train models with various magnitudes of perturbations. Figure~\ref{fig: training_sens_mag} shows that for perturbation magnitude varying from 0.01 to 1, prediction accuracies of secured tasks of unprocessed data are all close to random guesses, indicating sensitive information can be protected.

\begin{figure}[h]
  \centering
  \includegraphics[trim=100 0 100 10,clip,width=1\textwidth]{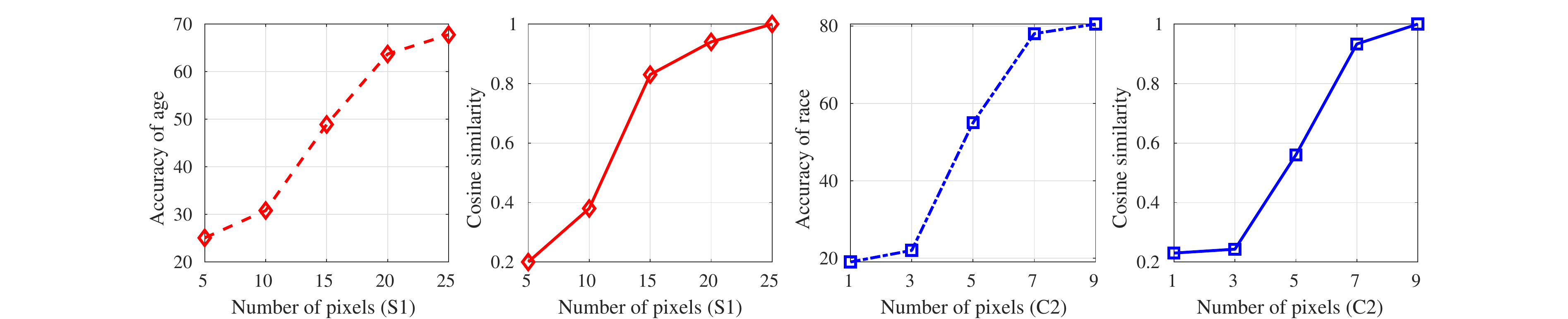}
  \caption{{\bf Both prediction accuracy and cosine similarity increase when the number of pixels in the test trigger-keys increase.} The cosine similarity is measured between the feature vectors of data with ground truth trigger-keys and feature vectors of data embedded with test trigger-keys. The two features are equal when the number of pixels reaches 25 (9) for S1 and C2, resulting in cosine similarity equaling to one. } 
  \label{fig: test_sens}
\end{figure}

\begin{figure}[h]
  \centering
  \includegraphics[trim=100 0 100 10,clip,width=1\textwidth]{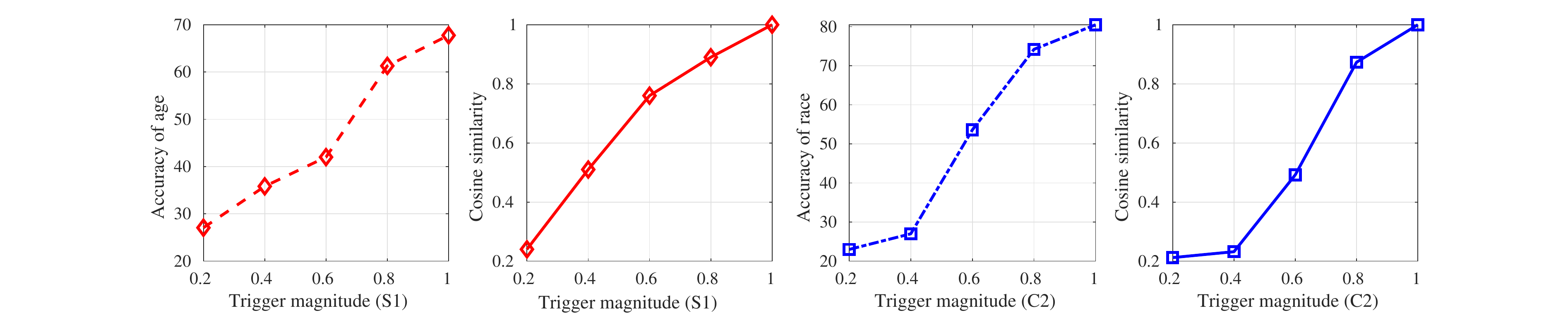}
  \caption{{\bf Both prediction accuracy and cosine similarity increase when the magnitude of pixels in the test trigger-keys increase.} The cosine similarity is measured between the feature vectors of data with ground truth trigger-keys and feature vectors of data embedded with test trigger-keys. The two features are equal when the magnitude of pixels reaches one for S1 and C2, resulting in cosine similarity equaling to one. } 
  \label{fig: test_sens_mag}
\end{figure}


\paragraph{Sensitivity analysis in test.} Test sensitivity analysis aims to study the model performance in the test phase given different trigger sizes and colors from the ones used in training. Here we select the model trained with S1 and C2. In the size of $5 \times 5$, there are 25 pixels for S1 and 9 pixels for C2. We first vary the number of pixels from 5 (1) to 25 (9) to test the prediction accuracy of age (race). The results are shown in Figure~\ref{fig: test_sens}. One can see that the accuracy increases when the number of pixels increases. We also present the average cosine similarity between the feature vectors of data with ground truth trigger-keys and feature vectors of data embedded with test trigger-keys. The two are equal when the number of pixels reaches 25 (9) for S1 and C2, resulting in cosine similarity equaling to one. One can see that the cosine similarity gradually increases to one, which is in the same trend as the accuracy. Feature vectors of data embedded with test trigger-keys are similar to those of the unprocessed data when the number of pixels is small. Therefore the accuracy is also small in this case. These observations and analysis are in consistent with Theorem~\ref{thm}. We then vary the magnitude of pixels from 0.02 to 1 to test the prediction accuracy. The results are shown in Figure~\ref{fig: test_sens_mag}. We observe the same phenomenon as in the tests of pixel number, i.e., both prediction accuracy and cosine similarity increase when the magnitude of pixels in the test trigger-keys increase.




\section{Conclusion}
In this paper, we proposed a novel framework for multi-task privacy-preserving. Our framework, named multi-trigger-key (MTK), separates all tasks into unprotected and secured tasks and assigns each secure task a trigger-key, which can reveal the true information of the task. Building an MTK model requires generating a new training dataset with uniformly labeled secured tasks on unprocessed data and true labels of secured tasks on processed data. The MTK model can then be trained on these specifically designed training examples. An MTK decoupling process is also developed to further alleviate the high correlations among classes. Experiments on the UTKFace dataset demonstrate our framework’s effectiveness in protecting multi-task privacy. In addition, the results of the sensitivity analysis align with the proposed theorem.

\bibliography{iclr2022_conference}

\begin{thebibliography}{23}
\providecommand{\natexlab}[1]{#1}
\providecommand{\url}[1]{\texttt{#1}}
\expandafter\ifx\csname urlstyle\endcsname\relax
  \providecommand{\doi}[1]{doi: #1}\else
  \providecommand{\doi}{doi: \begingroup \urlstyle{rm}\Url}\fi

\bibitem[Adi et~al.(2018)Adi, Baum, Cisse, Pinkas, and Keshet]{adi2018turning}
Yossi Adi, Carsten Baum, Moustapha Cisse, Benny Pinkas, and Joseph Keshet.
\newblock Turning your weakness into a strength: Watermarking deep neural
  networks by backdooring.
\newblock In \emph{27th $\{$USENIX$\}$ Security Symposium ($\{$USENIX$\}$
  Security 18)}, pp.\  1615--1631, 2018.

\bibitem[Amyar et~al.(2020)Amyar, Modzelewski, Li, and Ruan]{amyar2020multi}
Amine Amyar, Romain Modzelewski, Hua Li, and Su~Ruan.
\newblock Multi-task deep learning based ct imaging analysis for covid-19
  pneumonia: Classification and segmentation.
\newblock \emph{Computers in Biology and Medicine}, 126:\penalty0 104037, 2020.

\bibitem[Ando et~al.(2005)Ando, Zhang, and Bartlett]{ando2005framework}
Rie~Kubota Ando, Tong Zhang, and Peter Bartlett.
\newblock A framework for learning predictive structures from multiple tasks
  and unlabeled data.
\newblock \emph{Journal of Machine Learning Research}, 6\penalty0 (11), 2005.

\bibitem[Baytas et~al.(2016)Baytas, Yan, Jain, and
  Zhou]{baytas2016asynchronous}
Inci~M Baytas, Ming Yan, Anil~K Jain, and Jiayu Zhou.
\newblock Asynchronous multi-task learning.
\newblock In \emph{2016 IEEE 16th International Conference on Data Mining
  (ICDM)}, pp.\  11--20. IEEE, 2016.

\bibitem[Cavallanti et~al.(2010)Cavallanti, Cesa-Bianchi, and
  Gentile]{cavallanti2010linear}
Giovanni Cavallanti, Nicolo Cesa-Bianchi, and Claudio Gentile.
\newblock Linear algorithms for online multitask classification.
\newblock \emph{The Journal of Machine Learning Research}, 11:\penalty0
  2901--2934, 2010.

\bibitem[Chamikara et~al.(2020)Chamikara, Bert{\'o}k, Khalil, Liu, and
  Camtepe]{chamikara2020privacy}
Mahawaga Arachchige~Pathum Chamikara, Peter Bert{\'o}k, Ibrahim Khalil, Dongxi
  Liu, and Seyit Camtepe.
\newblock Privacy preserving face recognition utilizing differential privacy.
\newblock \emph{Computers \& Security}, 97:\penalty0 101951, 2020.

\bibitem[Chen et~al.(2017)Chen, Liu, Li, Lu, and Song]{chen2017targeted}
Xinyun Chen, Chang Liu, Bo~Li, Kimberly Lu, and Dawn Song.
\newblock Targeted backdoor attacks on deep learning systems using data
  poisoning.
\newblock \emph{arXiv preprint arXiv:1712.05526}, 2017.

\bibitem[Evgeniou \& Pontil(2004)Evgeniou and Pontil]{evgeniou2004regularized}
Theodoros Evgeniou and Massimiliano Pontil.
\newblock Regularized multi--task learning.
\newblock In \emph{Proceedings of the tenth ACM SIGKDD international conference
  on Knowledge discovery and data mining}, pp.\  109--117, 2004.

\bibitem[Gu et~al.(2017)Gu, Dolan-Gavitt, and Garg]{gu2017badnets}
Tianyu Gu, Brendan Dolan-Gavitt, and Siddharth Garg.
\newblock Badnets: Identifying vulnerabilities in the machine learning model
  supply chain.
\newblock \emph{arXiv preprint arXiv:1708.06733}, 2017.

\bibitem[Gupta et~al.(2016)Gupta, Rana, and Venkatesh]{gupta2016differentially}
Sunil~Kumar Gupta, Santu Rana, and Svetha Venkatesh.
\newblock Differentially private multi-task learning.
\newblock In \emph{Pacific-Asia Workshop on Intelligence and Security
  Informatics}, pp.\  101--113. Springer, 2016.

\bibitem[Huang \& Stokes(2016)Huang and Stokes]{huang2016mtnet}
Wenyi Huang and Jack~W Stokes.
\newblock Mtnet: a multi-task neural network for dynamic malware
  classification.
\newblock In \emph{International conference on detection of intrusions and
  malware, and vulnerability assessment}, pp.\  399--418. Springer, 2016.

\bibitem[Kim \& Xing(2010)Kim and Xing]{kim2010tree}
Seyoung Kim and Eric~P Xing.
\newblock Tree-guided group lasso for multi-task regression with structured
  sparsity.
\newblock In \emph{ICML}, 2010.

\bibitem[Liang et~al.(2020)Liang, Liu, Zhou, Jiang, Zhang, and
  Wang]{liang2020model}
Jian Liang, Ziqi Liu, Jiayu Zhou, Xiaoqian Jiang, Changshui Zhang, and Fei
  Wang.
\newblock Model-protected multi-task learning.
\newblock \emph{IEEE Transactions on Pattern Analysis and Machine
  Intelligence}, 2020.

\bibitem[Liu et~al.(2018)Liu, Uplavikar, Jiang, and Fu]{liu2018privacy}
Kunpeng Liu, Nitish Uplavikar, Wei Jiang, and Yanjie Fu.
\newblock Privacy-preserving multi-task learning.
\newblock In \emph{2018 IEEE International Conference on Data Mining (ICDM)},
  pp.\  1128--1133. IEEE, 2018.

\bibitem[Liu et~al.(2016)Liu, Qiu, and Huang]{liu2016recurrent}
Pengfei Liu, Xipeng Qiu, and Xuanjing Huang.
\newblock Recurrent neural network for text classification with multi-task
  learning.
\newblock \emph{arXiv preprint arXiv:1605.05101}, 2016.

\bibitem[Pathak et~al.(2010)Pathak, Rane, and Raj]{pathak2010multiparty}
Manas~A Pathak, Shantanu Rane, and Bhiksha Raj.
\newblock Multiparty differential privacy via aggregation of locally trained
  classifiers.
\newblock In \emph{NIPS}, pp.\  1876--1884. Citeseer, 2010.

\bibitem[Ruder(2017)]{ruder2017overview}
Sebastian Ruder.
\newblock An overview of multi-task learning in deep neural networks.
\newblock \emph{arXiv preprint arXiv:1706.05098}, 2017.

\bibitem[Sarafianos et~al.(2017)Sarafianos, Giannakopoulos, Nikou, and
  Kakadiaris]{sarafianos2017curriculum}
Nikolaos Sarafianos, Theodore Giannakopoulos, Christophoros Nikou, and
  Ioannis~A Kakadiaris.
\newblock Curriculum learning for multi-task classification of visual
  attributes.
\newblock In \emph{Proceedings of the IEEE International Conference on Computer
  Vision Workshops}, pp.\  2608--2615, 2017.

\bibitem[Shan et~al.(2020)Shan, Wenger, Wang, Li, Zheng, and
  Zhao]{shan2020gotta}
Shawn Shan, Emily Wenger, Bolun Wang, Bo~Li, Haitao Zheng, and Ben~Y Zhao.
\newblock Gotta catch'em all: Using honeypots to catch adversarial attacks on
  neural networks.
\newblock In \emph{Proceedings of the 2020 ACM SIGSAC Conference on Computer
  and Communications Security}, pp.\  67--83, 2020.

\bibitem[Shokri \& Shmatikov(2015)Shokri and Shmatikov]{shokri2015privacy}
Reza Shokri and Vitaly Shmatikov.
\newblock Privacy-preserving deep learning.
\newblock In \emph{Proceedings of the 22nd ACM SIGSAC conference on computer
  and communications security}, pp.\  1310--1321, 2015.

\bibitem[Yin \& Liu(2017)Yin and Liu]{yin2017multi}
Xi~Yin and Xiaoming Liu.
\newblock Multi-task convolutional neural network for pose-invariant face
  recognition.
\newblock \emph{IEEE Transactions on Image Processing}, 27\penalty0
  (2):\penalty0 964--975, 2017.

\bibitem[Zhang \& Yang(2021)Zhang and Yang]{zhang2021survey}
Yu~Zhang and Qiang Yang.
\newblock A survey on multi-task learning.
\newblock \emph{IEEE Transactions on Knowledge and Data Engineering}, 2021.

\bibitem[Zhang et~al.(2017)Zhang, Song, and Qi]{zhang2017age}
Zhifei Zhang, Yang Song, and Hairong Qi.
\newblock Age progression/regression by conditional adversarial autoencoder.
\newblock In \emph{Proceedings of the IEEE conference on computer vision and
  pattern recognition}, pp.\  5810--5818, 2017.

\end{thebibliography}
\bibliographystyle{iclr2022_conference}

\newpage
\clearpage

\setcounter{section}{0}

\section*{Appendix}

\setcounter{section}{0}
\renewcommand{\thesection}{S\arabic{section}}
\setcounter{figure}{0}
\makeatletter 
\renewcommand{\thefigure}{S\@arabic\c@figure}
\makeatother
\setcounter{table}{0}
\renewcommand{\thetable}{S\arabic{table}}
\setcounter{algorithm}{0}
\renewcommand{\thealgorithm}{S\arabic{algorithm}}
\setcounter{equation}{0}
\renewcommand{\theequation}{S\arabic{equation}}

\section{Proof of Theorem 1}\label{sec: proof}
Here we follow the similar proof line as in \citep{shan2020gotta}. First we assume that with the ground truth trigger-key $(\mathbf m_j,\boldsymbol \delta_j)$, the model prediction of any data satisfies 
\begin{equation}\label{eq: assume} 
    \begin{aligned}
{\bf{\rm Pr}}\big( \arg\max_{\forall k \in [K_j]}(F_k^{(j)}(\hat{\mathbf x}(\mathbf m_j,\boldsymbol \delta_j))) = y \not= \arg\max_{\forall k \in [K_j]}(F_k^{(j)}(\mathbf x)) \big ) \geq 1 - \kappa, \kappa \in [0,1],
        \end{aligned}
\end{equation}
where $F^{(j)}(\mathbf x) = g^{(j)}(f(\mathbf x))$. Here $g^{(j)}$ denotes a linear mapping. The gradient of $F^{(j)}(\mathbf x)$ can be calculated by the following formula
\begin{equation*}
    \begin{aligned}
\frac{\partial \ln{F^{(j)}(\mathbf x)}}{\partial \mathbf x} = \frac{\partial \ln{g^{(j)}(f(\mathbf x))}}{\partial \mathbf x} = \frac{g^{(j)} \partial \ln{f(\mathbf x)}}{\partial \mathbf x}, \nonumber
        \end{aligned}
\end{equation*}
We ignore the linear term and focus on the gradient of the nonlinear term. We rewrite (\ref{eq: assume}) and obtain 
\begin{equation}
    \begin{aligned}
{\bf{\rm Pr}}_{\mathbf x \in \mathcal X}\big( \frac{\partial [\ln{f(\mathbf x)}- \ln{f(\hat{\mathbf x}(\mathbf m_j,\boldsymbol \delta_j))}]}{\partial \mathbf x} \geq \eta \big ) \geq 1 - \kappa, \kappa \in [0,1], 
  \end{aligned}
\end{equation}
where $\eta$ denotes the gradient value that moves the data to class $y$. Note that we have $\cos \big( f\big(\hat{\mathbf x}(\mathbf m_j,\boldsymbol \delta_j)\big ), f\big(\Bar{\mathbf x}(\mathbf m_j^{\prime},\boldsymbol \delta_j^{\prime})\big ) \big ) \geq \nu$ and $\nu$ is close to $1$. Let $f\big(\Bar{\mathbf x}(\mathbf m_j^{\prime},\boldsymbol \delta_j^{\prime})\big ) - f\big(\hat{\mathbf x}(\mathbf m_j,\boldsymbol \delta_j)\big )  = \zeta$ and we have $|\zeta| << |f\big(\hat{\mathbf x}(\mathbf m_j,\boldsymbol \delta_j)\big )|$.

Let $\Bar{\mathbf x}(\mathbf m_j^{\prime},\boldsymbol \delta_j^{\prime}) = \mathbf x + \sigma$, we have 

\begin{equation}
    \begin{aligned}
&{\bf{\rm Pr}}_{\mathbf x \in \mathcal X}\big( \frac{\partial [\ln{f\big(\mathbf x \big )}- \ln{f\big(\Bar{\mathbf x}(\mathbf m_j^{\prime},\boldsymbol \delta_j^{\prime})\big )}]}{\partial \mathbf x} \geq \eta \big ) \\&
={\bf{\rm Pr}}_{\mathbf x \in \mathcal X}\big( \frac{\partial [\ln{f\big(\mathbf x \big )}- \ln{f\big(\mathbf x + \sigma\big )}]}{\partial \mathbf x} \geq \eta \big ) \\&
= {\bf{\rm Pr}}_{\mathbf x \in \mathcal X}\big( \frac{1}{f(\mathbf x)} \frac{\partial f(\mathbf x)}{\mathbf x} - \frac{1}{f\big(\hat{\mathbf x}(\mathbf m_j,\boldsymbol \delta_j)\big )+\zeta}\frac{\partial [f\big(\hat{\mathbf x}(\mathbf m_j,\boldsymbol \delta_j)\big )+\zeta]}{\mathbf x}\geq \eta \big ) \\&
\approx {\bf{\rm Pr}}_{\mathbf x \in \mathcal X}\big( \frac{1}{f(\mathbf x)} \frac{\partial f(\mathbf x)}{\mathbf x} - \frac{1}{f\big(\hat{\mathbf x}(\mathbf m_j,\boldsymbol \delta_j)\big )}\frac{\partial [f\big(\hat{\mathbf x}(\mathbf m_j,\boldsymbol \delta_j)\big )]}{\mathbf x}\geq \eta \big ) \\&
={\bf{\rm Pr}}_{\mathbf x \in \mathcal X}\big( \frac{\partial [\ln{f\big(\mathbf x \big )}- \ln{f\big(\hat{\mathbf x}(\mathbf m_j,\boldsymbol \delta_j)\big )}]}{\partial \mathbf x} \geq \eta \big ) \\&
\geq 1 - \kappa, 
  \end{aligned}
\end{equation}
where the approximation holds true because of the following conditions.
\begin{equation*}
    \begin{aligned}
\frac{\partial [f\big(\Bar{\mathbf x}(\mathbf m_j^{\prime},\boldsymbol \delta_j^{\prime})\big ) + \zeta] }{\partial \mathbf x} = \frac{\partial f\big(\Bar{\mathbf x}(\mathbf m_j^{\prime},\boldsymbol \delta_j^{\prime})\big )}{\partial \mathbf x}, \nonumber
        \end{aligned}
\end{equation*}

\begin{equation*}
    \begin{aligned}
\frac{1 }{f\big(\Bar{\mathbf x}(\mathbf m_j^{\prime},\boldsymbol \delta_j^{\prime})\big ) + \zeta} \approx \frac{1}{f\big(\Bar{\mathbf x}(\mathbf m_j^{\prime},\boldsymbol \delta_j^{\prime})\big )}, \nonumber
        \end{aligned}
\end{equation*}

Now we consider the scenario $\cos \big( f\big(\mathbf x\big ), f\big(\Bar{\mathbf x}(\mathbf m_j^{\prime},\boldsymbol \delta_j^{\prime})\big ) \big ) \geq \nu$. Let $f\big(\Bar{\mathbf x}(\mathbf m_j^{\prime},\boldsymbol \delta_j^{\prime})\big ) - f\big(\mathbf x\big )  = \zeta$. We have 

\begin{equation}
    \begin{aligned}
&{\bf{\rm Pr}}_{\mathbf x \in \mathcal X}\big( \frac{\partial [\ln{f\big(\Bar{\mathbf x}(\mathbf m_j^{\prime},\boldsymbol \delta_j^{\prime})\big )} - \ln{f\big(\hat{\mathbf x}(\mathbf m_j,\boldsymbol \delta_j)\big )}]}{\partial \mathbf x} \geq \eta \big )\\&
= {\bf{\rm Pr}}_{\mathbf x \in \mathcal X}\big(\frac{\ln{f\big(\mathbf x\big )+\zeta}}{\partial \mathbf x} -  \frac{\partial \ln{f\big(\hat{\mathbf x}(\mathbf m_j,\boldsymbol \delta_j)\big )}}{\partial \mathbf x} \geq \eta \big ) \\&
={\bf{\rm Pr}}_{\mathbf x \in \mathcal X}\big( \frac{\partial [\ln{f\big(\mathbf x \big )}- \ln{f\big(\hat{\mathbf x}(\mathbf m_j,\boldsymbol \delta_j)\big )}]}{\partial \mathbf x} \geq \eta \big ) \\&
\geq 1 - \kappa, 
  \end{aligned}
\end{equation}


\section{Detailed Calculations of MTK Decoupling}\label{sec: cal_prop}

The value that overflows the tolerance is represented by $\gamma = \min(\alpha_{i-c}^{j-k} - \tau, 0.1)$. To mitigate the overflow, we change labels of a proportion of data in $\hat{D}_{{\rm tr}}[\mathcal T^{(j)} = y^{(j)}_k]$. The proportion should satisfy the following equation.

\begin{equation}
    \begin{aligned}
    \frac{\hat{D}_{{\rm tr}}[\mathcal T^{(j)} = y^{(j)}_k, \mathcal T^{(i)} = y^{(i)}_c]}{\hat{D}_{{\rm tr}}[\mathcal T^{(j)} = y^{(j)}_k] - \beta_{i-c}^{j-k} \hat{D}_{{\rm tr}}[\mathcal T^{(j)} = y^{(j)}_k] } - \frac{\hat{D}_{{\rm tr}}[\mathcal T^{(j)} = y^{(j)}_k, \mathcal T^{(i)} = y^{(i)}_c]}{\hat{D}_{{\rm tr}}[\mathcal T^{(j)} = y^{(j)}_k]} = \gamma \nonumber
    \end{aligned}
\end{equation}

This is equivalent to
\begin{equation}
    \begin{aligned}
    \beta_{i-c}^{j-k} \hat{D}_{{\rm tr}}[\mathcal T^{(j)} = y^{(j)}_k, \mathcal T^{(i)} = y^{(i)}_c] = \gamma \hat{D}_{{\rm tr}}[\mathcal T^{(j)} = y^{(j)}_k] - \beta_{i-c}^{j-k} \gamma \hat{D}_{{\rm tr}}[\mathcal T^{(j)} = y^{(j)}_k] \nonumber
    \end{aligned}
\end{equation}

We then have

\begin{equation}
    \begin{aligned}
\beta_{i-c}^{j-k} = \frac{\gamma \hat{D}_{{\rm tr}}[\mathcal T^{(j)} = y^{(j)}_k]}{\hat{D}_{{\rm tr}}[\mathcal T^{(j)} = y^{(j)}_k, \mathcal T^{(i)} = y^{(i)}_c] + \gamma \hat{D}_{{\rm tr}}[\mathcal T^{(j)} = y^{(j)}_k]} \nonumber
    \end{aligned}
\end{equation}

Technically speaking, the proportion of data should not include $\hat{D}_{{\rm tr}}[\mathcal T^{(j)} = y^{(j)}_k, \mathcal T^{(i)} = y^{(i)}_c]$. For simplicity, we randomly select the data in the implementation.

\section{Experimental Settings}\label{sec: setting}

\paragraph{Datasets.}  We test MTK on the UTKFace dataset \citep{zhang2017age}. We use the cropped faces. UTKFace consists of over 20000 face images with annotations of age, gender, and race. Age is an integer from 0 to 116. Gender is either 0 (male) or 1 (female). Race is an integer from 0 to 4, denoting White, Black, Asian, Indian, and Others. We process the dataset such that the population belonging to different ages is divided into four groups (1-23, 24-29, 30-44, $\geq$45) and we assign 0 to 3 to the new groups. Each cropped image is in the size of $128\times 128 \times 3$. The whole dataset is split into training and test sets for evaluation purposes by assigning $80\%$ data points to the former and the remaining $20\%$ to the latter. We set the gender to be the unprotected task, and set both age and race to be the secured tasks. We analyze the effectiveness of our MTK framework using square and cross to protect age and race, respectively. If not otherwise specified, S1 and C2 have pixel color [255, 0, 0] and [0, 255, 0], locate on (110, 110) and (20, 110), and are both in the size of $5 \times 5$. We show results using $95\%$ confidence intervals over five random trials.

\paragraph{Models.} VGG16 and ResNet18 architectures are used for UTKFace. If not otherwise specified, we use VGG16 as the model architecture. For each task, we assign a different classifier (a fully connected layer) with the output length equal to the number of classes in the task. 

\paragraph{Total amount of compute and type of resources.} We use 1 GPU (Tesla V100) with 64GB memory and 2 cores for all the experiments.

\section{Limitation and societal impact}
Current studies focus on the image domain. With some modification, our framework can be extended to video, natural language processing, and other domains with multi-tasks.
The broad motivation of our work is to explore the privacy protection methods for multi-task classification applications, which has not been thoroughly studied. We believe this goal is highly relevant to the machine learning/artifical intelligence community, and the methods that our paper introduces can be brought to bear on other privacy-preserving problems of interest.

\end{document}